\documentclass[sigconf]{acmart}

\AtBeginDocument{%
  \providecommand\BibTeX{{%
    \normalfont B\kern-0.5em{\scshape i\kern-0.25em b}\kern-0.8em\TeX}}}

\usepackage{balance} 

\usepackage{graphicx}
\usepackage{amsmath}
\usepackage{makecell}
\usepackage{multirow}
\usepackage{amsthm}
\usepackage{booktabs}
\usepackage{algorithm}
\usepackage{algorithmic}
\usepackage{subfigure}
\usepackage{placeins}
\usepackage{xcolor}

\copyrightyear{2021}
\acmYear{2021}
\setcopyright{acmlicensed}\acmConference[CIKM '21]{Proceedings of the 30th ACM International Conference on Information and Knowledge Management}{November 1--5, 2021}{Virtual Event, QLD, Australia}
\acmBooktitle{Proceedings of the 30th ACM International Conference on Information and Knowledge Management (CIKM '21), November 1--5, 2021, Virtual Event, QLD, Australia}
\acmPrice{15.00}
\acmDOI{10.1145/3459637.3482115}
\acmISBN{978-1-4503-8446-9/21/11}

\begin{document}
\fancyhead{}

\title{Graph Embedding via Diffusion-Wavelets-Based Node Feature Distribution Characterization}

\author{Lili Wang}
\affiliation{%
  \institution{Dartmouth College}
  \city{Hanover}
  \state{New Hampshire}
  \country{USA}}
\email{lili.wang.gr@dartmouth.edu}

\author{Chenghan Huang}
\affiliation{%
  \institution{Millennium Management, LLC}
  \city{New York}
  \state{New York}
  \country{USA}}
\email{njhuangchenghan@gmail.com}

\author{Weicheng Ma}
\affiliation{%
  \institution{Dartmouth College}
  \city{Hanover}
  \state{New Hampshire}
  \country{USA}}
\email{weicheng.ma.gr@dartmouth.edu}

\author{Xinyuan Cao}
\affiliation{%
  \institution{Georgia Institute of Technology	}
  \city{Atlanta}
  \state{Georgia}
  \country{USA}}
\email{xcao78@gatech.edu}

\author{Soroush Vosoughi}
\affiliation{%
  \institution{Dartmouth College}
  \city{Hanover}
  \state{New Hampshire}
  \country{USA}}
\email{soroush.vosoughi@dartmouth.edu}

\begin{abstract}
Recent years have seen a rise in the development of representational learning methods for graph data. Most of these methods, however, focus on node-level representation learning at various scales (e.g., microscopic, mesoscopic, and macroscopic node embedding). In comparison, methods for representation learning on whole graphs are currently relatively sparse. 
In this paper, we propose a novel unsupervised whole graph embedding method. Our method uses spectral graph wavelets to capture topological similarities on each k-hop sub-graph between nodes and uses them to learn embeddings for the whole graph. We evaluate our method against 12 well-known baselines on 4 real-world datasets and show that our method achieves the best performance across all experiments, outperforming the current state-of-the-art by a considerable margin.

\footnotetext[1]{The first two authors contributed equally to this work.}

\end{abstract}



\maketitle
\linespread{1.01}
\section{Introduction}

Network data has become a ubiquitous part of daily life and spans diverse areas; from social networks, emails, and online forums, to scientific citation networks and protein or chemistry interactions. Accordingly, there has been a recent push to develop methods for knowledge extraction and representation learning for networks. When it comes to representation learning for graphs, the main area of focus has been node-level representation learning at different scales with relatively sparse attention given to methods for analyzing whole networks.

For instance, a fundamental problem in analyzing whole networks is to determine whether two graphs (or networks) are identical; this is also called the graph isomorphism problem. Babai \cite{babai2016graph} has shown that this problem can be solved in quasipolynomial time. In real-world applications, however, instead of determining whether two graphs are identical, we care about the similarity between graphs. A typical application of this approach involves classifying graphs based on their similarity. Note that this is a generalization of the graph isomorphism problem as two graphs that are identical will be labeled the same. One approach to solving the graph classification problem is to learn a representation of the graph as a vector, called whole graph embedding, which is invariant under the graph isomorphism and then adopt down-streaming classifiers.

In this paper, we propose a whole graph embedding method that considers node features as random variables and examines the distribution of node features in sub-graphs. 
Intuitively, the correlation between node features is related to the role similarity \cite{struc2vec} between them. For example, the nodes at centers of networks representing companies are likely to all be CEOs.
Based on this, we calculate the characteristic functions in k-hop sub-graphs and aggregate and sample the characteristic functions to generate the graph-level embedding. To capture topological similarity, we propose a diffusion-wavelet-based method. We then use the minimum difference of pair assignments (MDPA) \cite{MDPA}, a special case of earth mover's distance (EMD) \cite{rubner1998metric}, to measure the distance between the energy distributions for two nodes.

Specifically, we make the following contributions in this paper:
\vspace{-4.5pt}
\begin{itemize}

\item We present a framework for depicting the distribution of node features in sub-graphs based on diffusion wavelets and propose a graph-level embedding method based on the aggregation of characteristic functions.

\item We mathematically prove that our embedding method produces identical embeddings for isomorphic graphs. We further provide theoretical proof of the robustness of our method to feature noise.

\item We evaluate our method on the task of graph classification using four real-world networks. Our experiments show that our framework outperforms existing methods in learning whole graph representations.

\end{itemize}

\section{Related Work}
Many prior work have explored node representation learning \cite{node2vec, deepwalk, line, wang2021embedding, wang2020embedding, wang2021stress, wang2021hyperbolic,wang_tois_2021}. However, these methods do not work well on graph-level classification problems. The methods for graph classification can be grouped into several categories. A classic family of methods involve graph kernels with representative methods like the Weisfeiler-Lehman kernel \cite{shervashidze2011weisfeiler}, random walk kernel \cite{gartner2003graph}, shortest path kernel \cite{borgwardt2005shortest} and deep graph kernel \cite{yanardag2015deep}.  Another family of methods relies on graph embedding to learn a vector to represent a graph as a whole. Some of these methods are built upon graph kernels. For example, Graph2Vec \cite{narayanan2017graph2vec} first uses the Weisfeiler-Lehman kernel to extract rooted subgraph features which are then passed to a doc2vec \cite{le2014distributed} model to get embeddings. GL2Vec \cite{chen2019gl2vec} extends Graph2Vec by incorporating line graphs and in this way can deal with edge features. Other methods like SF \cite{de2018simple}, NetLSD \cite{tsitsulin2018netlsd}, and FGSD \cite{verma2017hunt} use the information from the Laplacian matrix and eigenvalues of graph to generate embeddings. Finally, Geo-Scatter \cite{gao2019geometric} and FEATHER \cite{feather} utilize the power of normalized adjacency matrices to capture the probability distribution of neighborhoods.

\section{Framework}
In this section, we formally introduce our framework. Let $G=(V, E, A)$ be an undirected and unweighted graph, where $V$ is a set of vertices, and $E \subseteq V \times V $ is the set of unweighted edges between vertices in $V$, and $A \in \mathbb{R}^{N \times m}$ describes the attributes of each node in the network. We consider the problem of representing the whole graph as one d-dimensional vector $X \in \mathbb{R}^d$, with $d<<|V|.$ Our framework combines the unique advantages of GraphWave \cite{graphwave} and FEATHER \cite{feather}, and consists of two parts: (1) topological wavelet similarity calculation and (2) sub-graph feature distribution characterization. We calculate node topological similarity based on diffusion wavelets, and we use that to capture the distribution of node features in sub-graphs. After aggregating the characteristic functions of k-hop sub-graphs, we pick representative sampling points and concatenate results to get the graph-level embedding. Below we describe these two parts in greater detail.
\subsection{Topological Wavelet Similarity}
\subsubsection{Diffusion Wavelets \cite{hammond2011wavelets}}
The Laplacian matrix $L$ is the difference between the adjacency matrix and the degree matrix of a graph. Assume $\lambda_1 \leq \lambda_2 \leq ...\leq \lambda_N$ are the eigenvalues of $L$, then $L$ can be decomposed as $L=U \Lambda U^T$, $\Lambda = diag(\lambda_1,...,\lambda_N)$. These eigenvalues describe the temporal frequencies of a signal on the graph. In order to discount larger eigenvalues and smooth the signals, a filter kernel $g_{\tau}$ with scaling parameter $\tau$ is introduced. Here, we use the heat kernel $g_{\tau} = e^{-\lambda{\tau}}$. The spectral wavelet coefficient matrix $\Psi$ is defined as:
\begin{equation}
\Psi = U\,diag(g_{\tau}(\lambda_1),...,g_{\tau}(\lambda_N)\,U^T
\end{equation}
For a given node $v_i$, the element $\Psi_{ji}$ represents how much energy comes from node $v_j$ to node $v_i$. Therefore, the $i$-th column of the wavelet coefficient matrix $\Psi$ describes a distribution of energy from the other nodes. It has been proved that nodes with similar energy distribution patterns have similar structural roles in the network \cite{graphwave}. Therefore, the difference between wavelet distributions of two nodes represents their topological distance.
\subsubsection{Topological Similarity}

The minimum difference of pair assignments (MDPA) can quickly measure the distance between two histograms\cite{MDPA}. It seeks the best one-to-one assignment between two lists to make the sum of the differences inside a pair to be minimized. Under certain conditions, the MDPA problem can be solved with linear time complexity. 
\begin{theorem}
Given two sets of \,$n$\, elements $X=(x_1, ..., x_n)$ and $Y=(y_1, ..., y_n)$, with $\forall i<j$, $x_i<x_j$ and $y_i<y_j$
. We must have the MDPA between $X$ and $Y$ as $\sum_{i=1}^n {|x_i-y_i|}$.
\end{theorem}
\begin{proof}
For any one-to-one assignment $((x_{i_1},y_{j_1}),...,(x_{i_n},y_{j_n}))$ between the $X$ and $Y$, if there exists $s$ and $t$ such that $x_{i_s}<x_{i_t}$ and $y_{j_s}>y_{j_t}$(or $x_{i_s}>x_{i_t}$ and $y_{j_s}<y_{j_t}$), we can always decrease the sum of the differences by switching $y_{j_s}$ and $y_{j_t}$. Therefore, the sum of differences achieves its minimum if and only if both $\{x_i\}$ and $\{y_j\}$ are ordered.
\end{proof}
We use the notation $\Psi_i$ for the spectral wavelet coefficients at a specific node $v_i$. In this way, to calculate the MDPA distance between $\Psi_i$ and $\Psi_j$, we just need to order both $\Psi_i$ and $\Psi_j$ to be ascending and calculate the pairwise distance. At last, after calculating the MDPA distance between pair of nodes $v_i$ and $v_j$, we define topological node similarity as follows:
\begin{equation}
s(v_i,v_j)=e^{-MDPA(\Psi_i,\Psi_j)}
\end{equation}
\subsection{Sub-graph Feature Distribution}
We assume that the features of node $v_i$ is a random vector $\hat{a}_i \in \mathbb{R}^m$, and $a_i$ in the attribute matrix $A$ can be considered as an observation. We are going to use the distribution of features in sub-graphs to recover the characteristic function of $\hat{a}_i$. Since the correlation between attributes are negatively related to the node distance \cite{cohen2014distance}, for a given node $v_i$, we consider the feature distribution in k-hop sub-graph $G_k(v_i)$. The characteristic function of $\hat{a}_i$ in $G_k(v_i)$ is
\begin{equation}
\label{eqn:charfunc}
\phi_{v_i}^{(k)}(t)=\mathbb{E}[e^{it{\hat{a}_i}}\vert G_k(v_i)]=\sum_{v_j \in G_k(v_i)}{\mathbb{P}(v_j|v_i)}e^{it{a_j}}
\end{equation}
The transition probability $\mathbb{P}(v_j|v_i)$ should be proportional to two factors: the similarity between nodes $v_j$ and $v_i$ and the influence of node $v_i$. We use normalized topological node similarity and normalized degree to calculate these values, respectively. The normalized topological node similarity is:
\begin{equation}
\label{eqn:normsimilarity}
\tilde{s}(v_i,v_j)=\frac {s(v_i,v_j)} {\sum_{v_r \in G_k(v_i)}{s(v_i,v_r)}}.
\end{equation}
Based on equation \ref{eqn:normsimilarity} and Euler's formula, we can expand equation \ref{eqn:charfunc} as:
\begin{equation}
\label{eqn:charfunc_euler}
\phi_{v_i}^{(k)}(t)=\sum_{v_j \in G_k}{\tilde{s}(v_i,v_j)(\cos(ta_j)+i\sin(ta_j))}.
\end{equation}
By aggregating the characteristic function over all nodes, we are then able to represent the graph level characteristic as:
\begin{equation}
\label{eqn:graphchar}
\phi_G^{(k)}(t)=\frac 1 {\lvert V \rvert}\sum_{v_i \in G}{\phi_{v_i}^{(k)}(t)}.
\end{equation}
We can sample equation \ref{eqn:graphchar} at $d$ evenly spaced points $t_1,...,t_d$ and concatenate them together to get the k-hop embedding:
\begin{equation}
\chi_{G_k}=[Re(\phi_G^{(k)}(t_i)),Im(\phi_G^{(k)}(t_i))]_{t_1,...,t_d}
\end{equation}
Concatenating the k-hop embeddings, we can get the graph level embedding based on topological similarity as:
\begin{equation}
\chi_G=\left[\chi_{G_1},\chi_{G_2},...,\chi_{G_{k_{max}}}\right]
\end{equation}
We can repeat this process to get the embedding with transition probability using normalized node influence. The final embedding $X$ is constructed by concatenating the embeddings with transition probability using normalized topological similarity and the embedding with transition probability using normalized node influence.\newline
\subsection{Theoretical Properties}
The following theorem shows that we can get the same embedding from isomorphic graphs by our method.
\begin{theorem}
Given two isomorphic graphs $G$ and $G'$, with the same sampling points $t_1, t_2, ..., t_d$, we have
\begin{equation}
\chi_G=\chi_{G'}
\end{equation}
\end{theorem}
\begin{proof}
\small
According to the definition of $\chi_G$, what we need to prove is $\forall k$,
\begin{equation}
\phi_G^{(k)}(t)=\phi_{G'}^{(k)}(t).
\end{equation}
We introduce a matrix $H^{(k)}$, where
\begin{equation}
H^{(k)}_{i,j}=\left\{
\begin{aligned}
& 1 \,\,\, & v_j \in G_k(v_i) \\
& 0 \,\,\, & v_j \not\in G_k(v_i)  \\
\end{aligned}\,\,\,.
\right.
\end{equation}
Apparently, $H^{(k)}$ is a symmetric matrix. We introduce another matrix $S$ with entries $S_{i,j}=\tilde{s}(v_i,v_j)$. Based on equation \ref{eqn:charfunc_euler} and equation \ref{eqn:graphchar},
\begin{equation}
\label{eqn:phi_G_k}
\begin{aligned}
\phi_G^{(k)}(t)&=\frac 1 {\lvert V \rvert}\sum_{v_i \in G}{\phi_{v_i}^{(k)}(t)}\\
&=\frac 1 {\lvert V \rvert}\sum_{v_i \in G}{\sum_{v_j \in G}{H^{(k)}_{i,j}S_{i,j}(\cos(ta_j)+i\sin(ta_j))}}
\end{aligned}
\end{equation}
$G$ and $G'$ are isomorphic, which means that there exists bijection $\Pi$ and $\pi$ such that
\begin{equation}
\begin{aligned}
&\Pi:G \to G' &v_i \mapsto v'_{i'},\\
&\pi:\{1,...,N\} \to \{1,...,N\} &i \mapsto i'.\\
\end{aligned}
\end{equation}
We use $H'$ and $S'$ to denote the $H$ and $S$ matrices for $G'$. In this way,
\begin{equation}
\label{eqn:phi_G_k}
\begin{aligned}
Re\left(\phi_{G'}^{(k)}(t)\right)&=\frac 1 {\lvert V \rvert}\sum_{v'_{i'} \in G'}{\sum_{v'_{j'}\in G'}{H'^{(k)}_{i',j'}S'_{i',j'}\cos(ta_{j'})}}\\
&=\frac 1 {\lvert V \rvert}\sum_{v_i \in G}{\sum_{v_j \in G}{H'^{(k)}_{\pi(i),\pi(j)}S'_{\pi(i),\pi(j)}\cos(ta_{\pi(j)})}}\\
&=\frac 1 {\lvert V \rvert}\sum_{v_i \in G}{\sum_{v_j \in G}{H^{(k)}_{i,j}S_{i,j}\cos(ta_j)}}\\
&=Re\left(\phi_{G}^{(k)}(t)\right)
\end{aligned}
\end{equation}
Similarly, $Im\left(\phi_{G'}^{(k)}(t)\right)=Im\left(\phi_{G}^{(k)}(t)\right)$.
\end{proof}
In addition to preserving the embedding for isomorphic graphs, another advantage of our method is its robustness against noisy features. This property is very useful in real-world scenarios. 
\begin{theorem}
Given undirected and attributed graph $G=(V,E,A)$ and its variant $G'=(V,E,A')$ with a noise in the features of $v_{j_0}$, for fixed sampling points $t_1, t_2, ..., t_d$, if $\left \Vert a_{j_0}-a'_{j_0} \right \Vert_{\infty} < \frac \epsilon {t_d}$ and $a_j=a'_j$ for any $j \neq {j_0}$, we must have $\left \Vert\chi_G-\chi_{G'}\right \Vert_2 < \epsilon$.
\end{theorem}
\begin{proof}
According to the definition of 2-norm, we have
\begin{equation}
\left \Vert\chi_G-\chi_{G'}\right \Vert_2 \leq \max_p\left\{\lvert \chi_G^{(p)}-\chi_{G'}^{(p)} \rvert \right\},
\end{equation}
where $\chi_G^{(p)}$ denotes the $p$-th component of $\chi_G$. We only have to prove that $\forall k$, $\forall p \leq m$ (m is the dimension of attributes), $\forall v_i \in V$,
\begin{equation}
\label{eqn:perturbation_re}
\lvert Re\left(\phi_{v_i}^{(k)}(t)\right)-Re\left(\phi_{v'_i}^{(k)}(t)\right)\rvert^{(p)} < \epsilon,
\end{equation}
\begin{equation}
\label{eqn:perturbation_im}
\lvert Im\left(\phi_{v_i}^{(k)}(t)\right)-Im\left(\phi_{v'_i}^{(k)}(t)\right)\rvert^{(p)} < \epsilon
\end{equation}
Here, we just show the proof of equation \ref{eqn:perturbation_re}, and the proof of equation \ref{eqn:perturbation_im} is similar. 
Plugging in equation \ref{eqn:charfunc_euler}, what we need to prove becomes
\begin{equation}
\label{eqn:eq_forref}
\lvert \sum_{v_j \in G_k}{\tilde{s}(v_i,v_j)\cos(ta_j^{(p)})}-\sum_{v_j' \in G'_k}{\tilde{s}(v_i',v_j')\cos(ta_j'^{(p)})}\rvert < \epsilon,
\end{equation}
Note that $G$ and $G'$ share the same topological structure, $\forall v_i,v_j \in V$,
\begin{equation}
\tilde{s}(v_i,v_j)=\tilde{s}(v_i',v_j'),
\end{equation}
where $v_i'$ and $v_j'$ are the corresponding nodes in $G'$. We have
\begin{equation}
\begin{aligned}
LHS\,\,of\,\,(\ref{eqn:eq_forref}) &=\lvert \tilde{s}(v_i,v_{j_0})\cos(ta_{j_0}^{(p)})-\tilde{s}(v_i',v'_{j_0})\cos(ta'\,\,^{(p)}_{j_0}) \rvert\\
&=\lvert \tilde{s}(v_i,v_{j_0})\left(\cos(ta_{j_0}^{(p)})-\cos(t{a'}_{j_0}^{(p)})\right) \rvert\\
&=\lvert \tilde{s}(v_i,v_{j_0})\sin(ta_{j_0}^{(p)}+\theta)t\left({a'}_{j_0}^{(p)}-a_{j_0}^{(p)}\right) \rvert\\
&\leq \lvert \tilde{s}(v_i,v_{j_0})\sin(ta_{j_0}^{(p)}+\theta)t\left \Vert a_{j_0}-a'_{j_0} \right \Vert_{\infty} \rvert\\
&< \epsilon
\end{aligned}
\end{equation}
\end{proof}
The same idea can be easily applied to the embedding with transition probability using normalized node influence.

\section{Experiment}
In this section, we use the classic task of graph classification to evaluate our method. We first introduce the detail of each dataset used in our experiment, then compare our method to 12 well-known baselines (including the current state-of-the-art, ``FEATHER'' \cite{feather}), and finally provide a parameter sensitivity analysis.
\subsection{Datasets}
We use four publicly available social graph datasets to evaluate our method. These datasets are from the Karate Club GitHub \cite{karateclub}: 

\begin{itemize}

\item \textbf{GitHub Repos}: This dataset consists of networks of developers who
starred GitHub repositories until August 2019. The nodes are Github users and the edges are follower relationships. The label of this dataset is whether a network belongs to web or machine learning developers.

\item \textbf{Reddit Threads}:  This dataset consists of networks of threads from Reddit collected in May 2018. The nodes are Reddit users and the edges are replies between them. The label of this dataset is whether a thread is discussion-based or not.

\item \textbf{Twitch Egos}:  This dataset consists of networks of ego-nets of users who participated in the partnership program in April 2018. The nodes are Twitch users and the edges are friendships. The label of this dataset is whether a user plays a single or multiple games.

\item \textbf{Deezer Egos}: This dataset consists of networks of ego-nets of users collected from the Deezer in February 2020. The nodes are Deezer users and edges are mutual follower relationships. The label of this dataset is the gender of the ego user.

\end{itemize}

Descriptive statistics of these datasets is shown in Table \ref{dataset}. Similar to prior work (e.g., Rozemberczki et al. \cite{feather}), for datasets without node features, we manually create two features for each node corresponding to the log degree and clustering coefficient of the node. 
\begin{table} 
\centering
\small
\begin{tabular}{cccccccc} 
\Xhline{2\arrayrulewidth}  
& & \multicolumn{2}{c} { Nodes } & \multicolumn{2}{c} { Density } & \multicolumn{2}{c} { Diameter } \\
\Xhline{2\arrayrulewidth}  
Dataset & Graphs & Min & Max & Min & Max & Min & Max \\
  \Xhline{2\arrayrulewidth} 
  GitHub Repos & 12,725 & 10 & 957 & $0.003$ & $0.561$ & 2 & 18 \\
  Reddit Threads & 203,088 & 11 & 97 & $0.021$ & $0.382$ & 2 & 27 \\
Twitch Egos & 127,094 & 14 & 52 & $0.038$ & $0.967$ & 1 & 2 \\

Deezer Egos & 9,629 & 11 & 363 & $0.015$ & $0.909$ & 2 & 2 \\
  \Xhline{2\arrayrulewidth}
\end{tabular}
\caption{Descriptive statistics of the four datasets used in the graph classification experiments. This table is taken from Rozemberczki et al. \cite{feather}.}  
\label{dataset}
\end{table}
\subsection{Graph classification}
To better compare our model to prior work, we use the same settings specified by Rozemberczki et al. \cite{feather}: We run the graph classification task on each dataset with ten $80 \% / 20 \%$ train-test split ratios from different random seeds (from 0 to 9) and use an off-the-shelf logistic regression model (using scikit-learn), with the default parameters and the SAGA optimizer for classification. The averaged AUC scores with corresponding standard errors are reported. For all the baselines -- GL2Vec, Graph2Vec, SF, NetLSD, FGSD, Geo-Scatter, FEATHER, Mean Pool, Max Pool, Sort Pool \cite{zhang2018end}, Top K Pool \cite{gao2019graph}, and SAG Pool \cite{lee2019self}, we show the results reported by Rozemberczki et al. \cite{feather}. As shown in Table \ref{graphclassification}, our method outperforms all the baselines on all the datasets.

\begin{table} 
\centering

\begin{tabular}{lllll}
\Xhline{2\arrayrulewidth}
     & \thead{GitHub\\Repos} & \thead{Reddit\\Threads} & \thead{Twitch\\Egos} & \thead{Deezer\\Egos}  \\
     \Xhline{2\arrayrulewidth}

    GL2Vec &  .532$\pm$.002 & .754$\pm$.001 & .670$\pm$.001& .500$\pm$.001 \\ 
    Graph2Vec  &  .563$\pm$.002 & .808$\pm$.001 & .698$\pm$.001& .510$\pm$.001 \\ 
    SF  &  .535$\pm$.001 & .819$\pm$.001 & .642$\pm$.001& .503$\pm$.001 \\ 
    NetLSD  &  .614$\pm$.002 & .817$\pm$.001 & .630$\pm$.001& .525$\pm$.001 \\ 
    FGSD  &  .650$\pm$.002 & .822$\pm$.001 & .699$\pm$.001& .528$\pm$.001 \\ 
    Geo-Scatter  &  .532$\pm$.001 & .800$\pm$.001 & .695$\pm$.001& .524$\pm$.001 \\ 
     FEATHER  &  .728$\pm$.002 & .823$\pm$.001 & .719$\pm$.001& .526$\pm$.001 \\ 
    Mean Pool  &  .599$\pm$.003 & .801$\pm$.002 & .708$\pm$.001& .503$\pm$.001 \\ 
    Max Pool  &  .612$\pm$.013 & .805$\pm$.001 & .713$\pm$.001& .515$\pm$.001 \\ 
    Sort Pool  &  .614$\pm$.010 & .807$\pm$.001 & .712$\pm$.001& .528$\pm$.001 \\ 
    Top K Pool  &  .634$\pm$.001 & .807$\pm$.001 & .706$\pm$.002& .520$\pm$.003 \\ 
    SAG Pool   &  .620$\pm$.001 & .804$\pm$.001 & .705$\pm$.002& .518$\pm$.003 \\

    Our method & \textbf{.772$\pm$.002}  &  \textbf{.835$\pm$.001} &  \textbf{.722$\pm$.001} & \textbf{.538$\pm$.003} \\ 
\Xhline{2\arrayrulewidth}
\end{tabular}
  \caption{Average AUC scores (and standard errors) of our model vs all the baselines for the graph classification task. The baseline results are taken from Rozemberczki et al. \cite{feather}.}  
   \label{graphclassification}

\end{table}

\subsection{Parameter Sensitivity Analysis}
\begin{figure}[htbp]
\centerline{\includegraphics[width=1.05\linewidth]{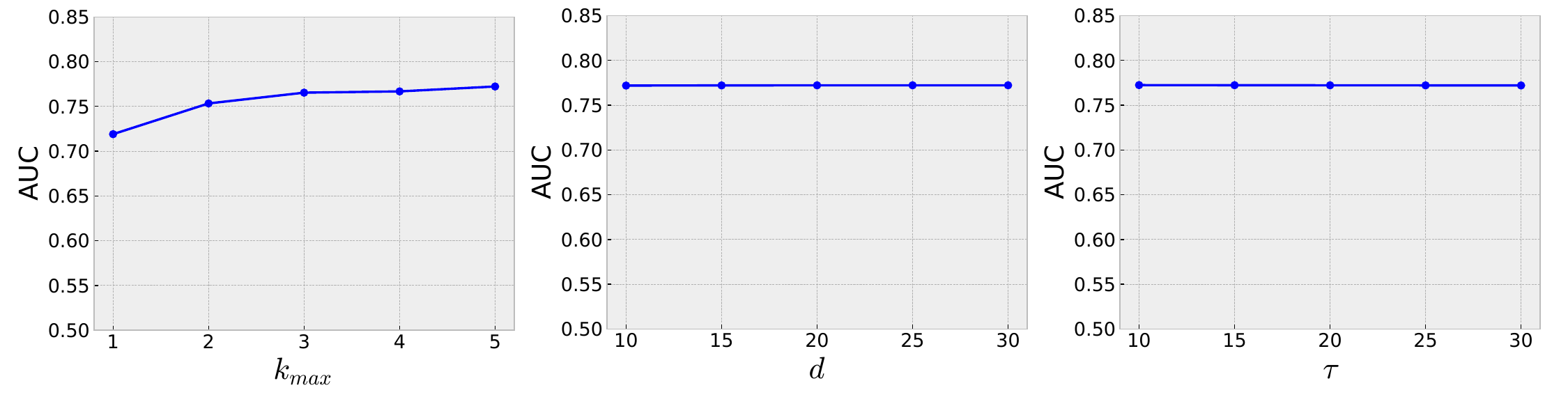}}
\caption{Parameter sensitivity for the graph classification task on the GitHub Repos dataset.}
\label{para}
\end{figure}
In this section, we study the sensitivity of our method to the choice of hyper-parameters. The hyper-parameters used in our model are:

\begin{itemize}

\item \textbf{$k_{max}$ -} (default: 5) The maximum scale of k-hop sub-graph to capture the topological similarity.

\item \textbf{$d$ -} (default: 25) The number of sampling points.
 
\item \textbf{$\tau$ -} (default: 0.5) The scaling parameter of the filter kernel.

\end{itemize}

The parameter sensitivity analysis for the graph classification task on the GitHub Repos dataset is shown in Fig. \ref{para}. We tune each parameter separately while fixing the other parameters to the default value. Overall, our model is not parameter sensitive. The results are stable for all the different number of sampling points ($d$) and $\tau$, and all the scales ($k_{max}$) larger than two.

\section{Conclusion}
In this paper, we introduced a novel framework to depict the distribution of node features in k-hop sub-graphs based on diffusion wavelets and proposed a graph-level embedding method based on the aggregation of the characteristic functions. We also provide theoretical proofs that our embedding method produces identical embeddings for isomorphic graphs and that it is robust to feature noise. We evaluated our method on the task of graph classification using four real-world networks and compared it against 12 baselines. Our method outperformed them all in all of our experiments, achieving the current state-of-the-art. 

\noindent \paragraph{Code \& Data Availability:}The code and data for this paper will be made available upon request.

\bibliographystyle{ACM-Reference-Format}
\balance

\bibliography{sample-base}

\end{document}